\RequirePackage{fix-cm}
\documentclass{svjour3}                     
\usepackage{graphicx}
\usepackage[utf8]{inputenc} 
\usepackage{url}            
\usepackage{booktabs}       
\usepackage{nicefrac}       
\usepackage{microtype}      
\usepackage{amsmath,amssymb,amsfonts}
\usepackage{multirow}
\usepackage{times}
\usepackage{latexsym}
\usepackage{array}
\usepackage{bm}
\usepackage{ulem}
\usepackage{algorithm}
\usepackage[noend]{algorithmic}
\usepackage{balance}
\usepackage{subfigure}
\usepackage{diagbox} 
\usepackage{float}
\usepackage{graphicx}
\usepackage{textcomp}
\usepackage{xcolor}

\newcommand{\comment}[1]{}

\newcommand\oursstl {\textbf{SDA}}
\newcommand\ourssfe {\textbf{TOE}}

\begin{document}
\title{Unsupervised Sentiment Analysis \\by Transferring Multi-source Knowledge
}


\author{Yong Dai         \and
        Jian Liu    \and
        Jian Zhang    \and
        Hongguang Fu \and
        Zenglin Xu 
}

\institute{Yong Dai, Jian Liu, Jian Zhang, Hongguang Fu \at
              SMILE Lab, School of Computer Science and Engineering, University of Electronic Science and Technology of China, Chengdu, Sichuan, China
           \and
          Zenglin Xu (Corresponding author)\at
           School of Computer Science and Technology, Harbin Institute of Technology, Shenzhen, China 
            \email{xuzenglin@hit.edu.cn}
}

\date{Received: date / Accepted: date}

\maketitle


\begin{abstract}

\textbf{Background}. Sentiment analysis (SA) is an important research area in cognitive computation—thus in-depth studies of patterns of sentiment analysis are necessary. At present, rich-resource data-based SA has been well-developed, while the more challenging and practical multi-source unsupervised SA (i.e. a target-domain SA by transferring from multiple source domains) is seldom studied. The challenges behind this problem mainly locate in the lack of supervision information, the semantic gaps among domains (i.e., domain shifts), and the loss of knowledge. However, existing methods either lack the distinguishable capacity of the semantic gaps among domains or lose private knowledge. 

\textbf{Methods.}
To alleviate these problems, we propose a two-stage domain adaptation framework. In the first stage, a multi-task methodology-based shared-private architecture is employed to explicitly model the domain-common features and the domain-specific features for the labeled source domains. In the second stage, two elaborate mechanisms are embedded in the shared-private architecture to transfer knowledge from multiple source domains. The first mechanism is a selective domain adaptation (\oursstl) method, which transfers knowledge from the closest source domain. And the second mechanism is a target-oriented ensemble (\ourssfe) method, in which knowledge is transferred through a well-designed ensemble method.

\textbf{Results and Conclusions}.
Extensive experiment evaluations verify that the performance of the proposed framework outperforms unsupervised state-of-the-art competitors. What can be concluded from the experiments is that transferring from very different distributed source domains may degrade the target-domain performance, and it is crucial to choose proper source domains to transfer from.

\end{abstract}

\section{Introduction}
Cognitive computing is a process of mimicking the functioning of the human brain to help improve human decision-making from the unstructured data, especially the natural language. Sentiment analysis (SA) has been an active research topic in natural language processing, which is a cognitive computing study of people’s opinions, sentiments, emotions, appraisals, and attitudes towards entities such as products, services, organizations, individuals, issues, events, topics, and their attributes~\cite{Zhang2018DeepLF}. SA is of academic and industrial importance and value in areas like e-commerce (e.g., recommendation systems), education, opinion polls, user behavior modeling, etc.~\cite{quan2014unsupervised,nguyen2015sentiment,mantyla2018evolution,Zhang2018DeepLF,liu2019cikm}. Generally, sentiment classification is domain-dependent~\cite{blitzer2007biographies}, that is to say, the sentiment polarity of a comment may be contradicting because of the semantic gaps between domains. For example, a comment 'It runs slow' expresses a positive attitude to 'battery', but negative to 'motorcycle'; alternatively, some comments may have the same polarities for most domains, such as the comment 'It is wonderful'. Therefore, comments from diverse products or services must be sampled from different data distributions, and the model trained for one domain cannot be directly applied to another domain. 

Nevertheless, the real situations are: (1) there often exist many domains of comments in the meanwhile; and (2) there is an adequate amount of labeled training data for every domain of interest is typically impractical, which makes the further study on how to handle with the limited-resource multiple domains of data necessary and worthwhile. To this end, a series of powerful multi-task models, which consider the system learning for each domain as an independent task so that each task can reinforce and complement each other, is established~\cite{wu2015collaborative,liu2016recurrent,liu2017adversarial,chen2018multinomial,chen2018zero,gholami2018unsupervised,zhang2018shaped,zheng2018same,collobert2008unified,liu2016deep,liu2016recurrent,ruder2017overview,LiuXDBFRZ17,XiongSYQHHL19}. Among them, a group of representative studies employ a shared-private model~\cite{liu2016deep,wu2015collaborative,liu2016recurrent,liu2017adversarial,chen2018multinomial,chen2018zero,gholami2018unsupervised,zhang2018shaped,zheng2018same}, which introduces two feature spaces for any task: one is used to store task-dependent features, the other is used to capture shared features. In the early works~\cite{wu2015collaborative,liu2016recurrent,liu2016deep}, researchers divide the features of different domains into private and shared spaces merely based on whether features from different domains should be shared. The major limitation of these methods is that the shared feature space could contain some unnecessary domain-specific features, while some domain-invariant features could also be mixed in private spaces, which would degrade the performance of the multi-task systems.

To alleviate the shared and private latent feature spaces from interfering with each other, the domain adversarial training was first integrated into the multi-task system~\cite{liu2017adversarial}. The domain adversarial training originally aims at learning latent feature representations that serve at reducing the discrepancy between the source and target distribution by minimising the discrepancies between training and synthetic data distributions~\cite{goodfellow2014generative}. The theoretical ground behind the adversarial training is that cross-domain generalisation can be achieved by means of feature representations for which the domain of the input example cannot be identified~\cite{ben2010theory}. Then, a multinomial adversarial network (MAN)~\cite{chen2018multinomial} was proposed, which provides theoretical justifications proving that their methods are essentially minimizers of various f-divergence~\cite{ali1966general} among multiple probability distributions. Although these methods can model the different domains of data well, they have no ability to extract private features when handling the domain with no labeled data. However, there may always have an amount of the emerging domains of comments (e.g. Amazon customers comment on new products) with no labeled data. On the other hand, the most studied one-to-one domain adaptation methods cannot be applied directly to transfer knowledge from multiple source domains~\cite{sun2015survey}.

In this paper, we specialize in the multi-source unsupervised domain adaptation (MS-UDA) for sentiment analysis, where there are multiple domains with labeled data (i.e. source domains) and one domain of interest with unlabeled data (i.e. target domain). To better deal with this setting, we apply the shared-private methodology to the MS-UDA setting and propose a two-stage multi-task learning model by embedding two elaborately designed mechanisms into the learning system, through which we can not only retain the good modeling capability of multiple domains of the data but also mitigate the loss of the private knowledge. In the first stage, we pre-train an adversarial shared-private model for all source domains aiming at modeling the shared and private features explicitly in a supervised way and obtaining a good decision boundary for each source domain. In the second stage, we introduce two domain adaptation mechanisms to transfer domain-specific knowledge from the source domain(s) at the feature level and classifier level separately to make up the knowledge loss for the target domain. In detail, the first mechanism is a selective domain adaptation (\oursstl)~\cite{DaiZYX19} mechanism, which adapts the knowledge from the closest source domain by introducing another discriminator to align the private feature distributions between the target domain and the selected source domain. By \oursstl, we turn the MS-UDA problem into the traditional one-to-one domain adaptation problem. The second mechanism is a target-oriented ensemble (\ourssfe) method, which is inspired by the Tri-training~\cite{zhou2005tri} method and the ensemble learning~\cite{zhang2012ensemble,polikar2012ensemble} methods. In this mechanism, we select the top-3 closest source domains, annotate unlabeled target instances by them, finetune the selected source extractors using these annotated instances, and at last make inference via the finetuned source models at the classifier level. 

Our contributions are summarized as follows:
\begin{enumerate}
\itemsep=0pt
\item We propose a two-stage multi-task learning-based framework with two levels of knowledge transferring mechanisms for sentiment analysis in the multi-source unsupervised domain adaptation setting. The proposed framework can model the shared and private knowledge well for the labeled source domains and the unlabeled target domain. Moreover, this framework can be extended to the multi-source multi-target setting.
\item Our proposed framework outperforms the state-of-the-art unsupervised domain adaptation methods, and even works better on some certain domains than the state-of-the-art supervised models.
\end{enumerate} 
\section{Related Work}
\label{related}
\subsection{Cross-domain sentiment analysis}
In the real scenario, there often exists a number of data for SA with different distributions (i.e. multiple domains of data) at the same time and it is not practical to annotate an abundance of data for every existing and upcoming domain. But a model trained for one domain may not perform well in another domain due to the distribution discrepancy. So the cross-domain SA, which transfers knowledge from the rich-source domains (denoted as source domains) to low-source domains (denoted as target domains), is widely explored by different lines of work.

One line of literature focus on the one-to-one cross-domain SA~\cite{blitzer2007biographies,pan2010cross,li2009knowledge,he2011automatically,gezici2015sentiment,mao2015cross}. The structural correspondence learning (SCL) algorithm~\cite{blitzer2007biographies} implement domain adaptation at the feature level based on the selected pivot features. The spectral feature alignment (SFA) algorithm~\cite{pan2010cross} attempts to align the domain-specific sentiment words from different domains into clusters. In~\cite{gezici2015sentiment}, authors adapt the sentiment scores of a general-purpose sentiment lexicon to a specific domain. In~\cite{mao2015cross}, the authors depend on the construction of domain-specific lexicons to improve cross-domain sentiment analysis. Despite the evolving approaches, the performance will degrade obviously when dealing with a big distribution gap between source and target domain. More importantly, they are intuitively unable to exploit knowledge from multiple source domains.

Another line of research explores to transfer knowledge from multiple domains~\cite{glorot2011domain,duan2009domain,wu2016sentiment,ash2016unsupervised,Fang18BGSHXcorr}. Stacked Denoising Auto-encoders (SDA) extracts a high-level representation to capture the common concepts from multiple source domains~\cite{glorot2011domain}. The Domain Adaptation Machine (DAM) method learns a Least-Squares SVM classifier for the target domain by leveraging the classifiers independently trained in multiple source domains~\cite{duan2009domain}. In~\cite{wu2016sentiment}, the authors proposed to extract domain-specific and global sentiment-related information from multiple source domains. In~\cite{ash2016unsupervised}, the authors proposed an unsupervised multi-source domain adaptation methodology based on an assumption that source and target domains have similar feature distributions. This kind of methods either adopt the unreasonable assumptions, cannot utilize the unlabeled data, or base on the handcrafted features.

 Our work is motivated by recent advances in  multiview learning~\cite{HuangRX18,HuangKX18,HuangKTX19,DBLP:journals/isci/HuangXTK20}, multiple kernel learning~\cite{poria-etal-2015-deep,KangWCX19}, tensor analysis~\cite{zadeh-etal-2017-tensor,LiuXDBFRZ17}, graph learning~\cite{kang2020relation}, and adversarial learning~\cite{YouYLX019,LiangCZCBX19}. Compared to previous works on multi-source sentiment analysis, our proposed framework focuses on transferring knowledge from multi-source domains with an ability of effective utilization of unlabeled data in an end-to-end way.
\subsection{Multi-task learning}
The key issue of multi-task learning is to train a well-performed model by leveraging the common characteristic between tasks or domains~\cite{caruana1997multitask,ruder2017overview,bousmalis2016domain,liu2017adversarial,chen2018multinomial}. The hard parameter sharing method is generally applied by sharing the hidden layers between all tasks while keeping several task-specific output layers~\cite{caruana1997multitask}. Due to the sharing of the feature encoder, the hard-shared architecture cannot extract the domain-specific knowledge. In~\cite{bousmalis2016domain,liu2017adversarial,chen2018multinomial}, the authors adopted shared-private architecture to perform multi-source SA, in which shared and private features are efficiently used on the condition all domains have labeled data. In~\cite{liu2017adversarial}, the domain adversarial training is first introduced to the multi-task learning to extract the domain-shared features. Next, the domain adversarial training is theoretical grounded and extended to the multi-source domain handling~\cite{chen2018multinomial}. But they all lose the private knowledge when dealing with the unlabeled target domain. In~\cite{zheng2018same}, they utilize shared sentence encoders but private query features to select domain-specific information from shared sentence representation but ignores identifying sentiment polarity of the same word in different domains. 

In our paper, we follow the shared-private methodology to introduce a framework with an ability to model the shared and private features reasonably for the labeled source domains and further embed two elaborated mechanisms into this framework to make up the lost knowledge for the unlabeled target domain.

\section{Methodology}
In this section, we introduce the proposed framework for unsupervised sentiment analysis with multiple source domains. Our framework includes two stages: the first stage is training an adversarial shared-private model for all source domains with supervision information; the second stage is to transfer knowledge from multiple source domains through two mechanisms at either the feature level or the classifier level. We first present the problem definition and notations (\ref{notations}), followed by an overview of the architecture adopted (\ref{overview}). Then we detail the two mechanisms employed in the second stage (\ref{fist} and \ref{second}). 

\subsection{Problem Definition and Notations}\label{notations}
Suppose we have $K$ labeled source domains $\left \{ \mathcal{S} _j\right \}_{j=1}^{K}$ and one unlabeled target domain $\mathcal{T}$, where $
    \mathcal{S}_j  \triangleq \left ( X_{S_j},Y_{S_j} \right )\triangleq \left \{ (\boldsymbol{x}_{i}^{\mathcal{S}_j},\boldsymbol{y}_{i}^{\mathcal{S}_j}) \right \}_{i=1}^{\left | \mathcal{S}_j \right |}$ and
    $\mathcal{T} \triangleq \left ( X_t \right ) \triangleq \left \{ \boldsymbol{x}_{i}^{\mathcal{T}} \right \}_{i=1}^{\left | \mathcal{T} \right |}$.
 $\boldsymbol{x}_{i}^{\mathcal{S}_j}$ and $\boldsymbol{y}^{\mathcal{S}_j}$ are comment sentences and their corresponding sentiment labels (i.e. positive or negative) coming from the $j$-th ($1\leq j\leq K$) source domain and $\boldsymbol{x}_{i}^{\mathcal{T}}$ denotes examples from the target domain. The multi-source unsupervised domain adaptation (MS-UDA) aims to learn a function $f$ from multiple source domains (i.e. $\left \{ \mathcal{S} _j\right \}_{j=1}^{K}$) that generalize well to the unlabeled target domain (i.e. $\mathcal{T}$). For convenience, we denote $N_s=\sum_{j=1}^{K}\left | S_j \right |$, $N_\mathcal{T}=\left | \mathcal{T} \right |$ as the size of all labeled source data and unlabeled target data respectively,  and $\mathcal{U} \triangleq \left \{ (\boldsymbol{x}_{i},\boldsymbol{d}_{i}) \right \}_{i=1}^{N}$ as all the data with domain information $d_i$, where $N=N_s+N_\mathcal{T}$.

\subsection{An overview of the adversarial shared-private architecture}\label{overview}
\begin{figure}
  \centering
  \centerline{\includegraphics[scale=0.3]{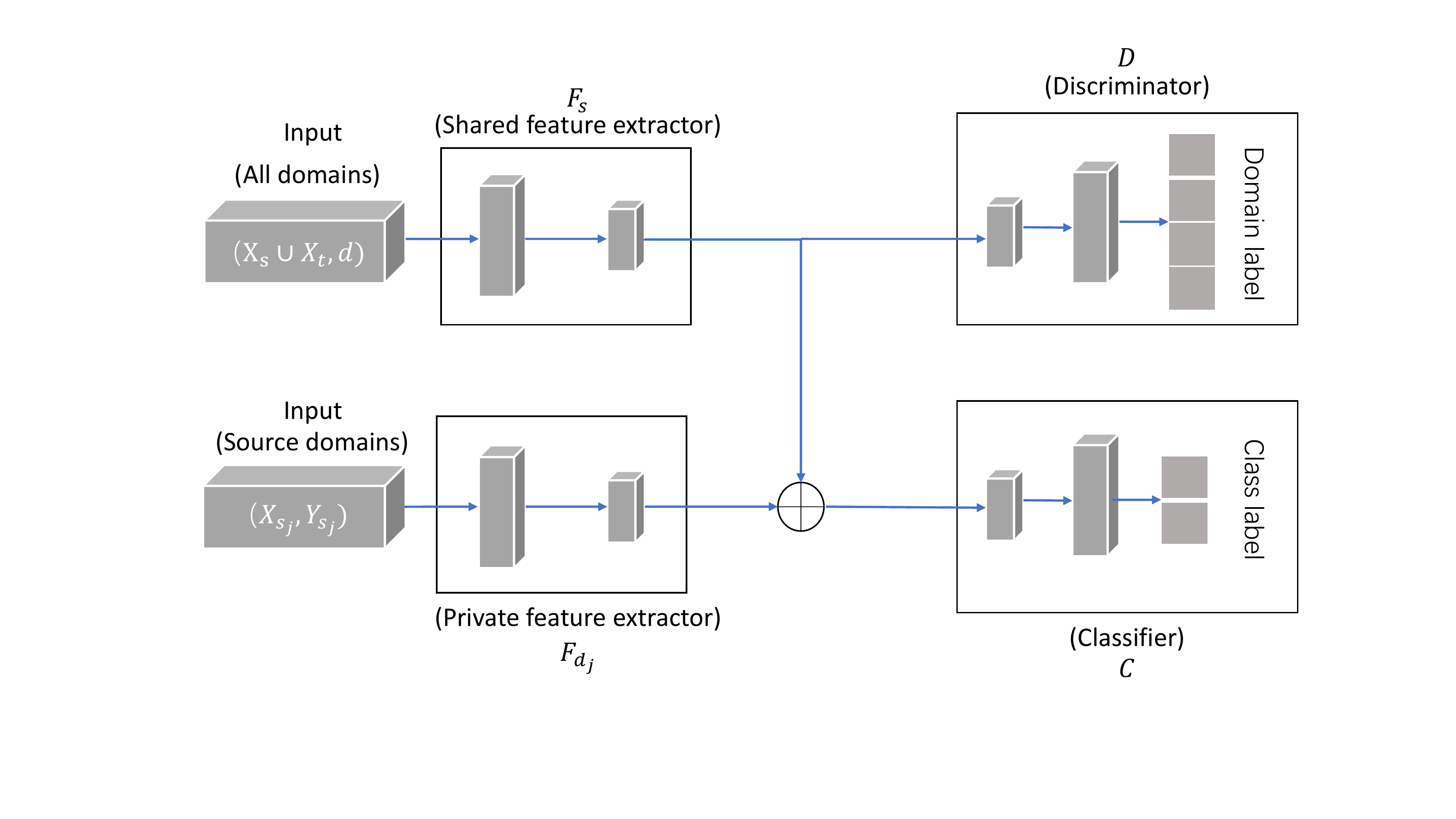}}
  \caption{The adversarial shared-private architecture. The cubes in the first column are inputs from different domains. $\mathcal{F}_s$ and $\left \{ \mathcal{F}_{d_j} \right \}_{j=1}^{K}$ in the second column denote the shared feature extractor for all domains and private feature extractor for each source domain, respectively.  $\mathcal{D}$ denotes a discriminator for adversarial training, and  $\mathcal{C}$  denotes a classifier for sentiment classification. Shared features extracted by $\mathcal{F}_s$ and constrained by $D$ are concatenated with private features extracted by $\mathcal{F}_{d_j}$ from each domain, and then the concatenated features are sent to classifier $\mathcal{C}$ to conduct the sentiment classification.}
  \label{step1}
\end{figure}
In this subsection, we present the adversarial shared-private multi-task learning architecture, which is adopted in~\cite{liu2017adversarial,chen2018multinomial} and employed in our paper to build a well-behaved model for all the source domains and form a sound transferring basis for the target domain. As Figure \ref{step1} depicts, the basic architecture includes a shared feature extractor $\mathcal{F}_s$, a private feature extractor $\mathcal{F}_{d_j}$ for the $j$th domain, a text classifier $\mathcal{C}$, and a domain discriminator $\mathcal{D}$. Each module in this architecture can be flexibly decided according to the different situations. 

As the name implies, the shared-private architecture separates features into the shared part (shared across the domains) and the private part (owned by each domain itself). To derive purely shared features, we assume the existence of a shared feature space between domains where the distribution divergence is small, and reducing domain divergence through the adversarial training can exploit domain invariant features (i.e. shared features)~\cite{ben2010theory}. In practice, a discriminator $D$ is utilized to implement the adversarial training~\cite{ganin2014unsupervised,ganin2016domain,liu2017adversarial,chen2018multinomial}. The objective function can be written as:
\begin{equation}\label{eqa:d}
    J_{\mathcal{D}}=\sum_{j=1}^{K}\sum_{i=1}^{\left | \mathcal{S}_j \right |} \underset{x_i \sim \left \{ \mathcal{S} _j\right \}}{\mathbb{E}}\left[\mathcal{L}_{\mathcal{D}}\left(\mathcal{D}\left(\mathcal{F}_{s}(x_i)\right) ; d_i\right)\right],
\end{equation}
where $d_i$ is the domain label for each instance and $\mathcal{L}_{\mathcal{D}}$ is a loss function (e.g., the canonical negative log-likelihood (NLL)). After convergence of the adversarial training, we concatenate the shared features with the private features and send them into the classifier $C$ to fulfill the classification. The objective of $C$ is defined as:
\begin{equation}
    J_{\mathcal{C}}=\sum_{j=1}^{K}\sum_{i=1}^{\left | \mathcal{S}_j \right |} \underset{(x_i,y_i) \sim \left \{ \mathcal{S} _j\right \}}{\mathbb{E}}\left[\mathcal{L}_{\mathcal{C}}\left(\mathcal{C}\left(\mathcal{F}_{s}(x_i), \mathcal{F}_{j}(x_i)\right) ; y_i\right)\right],
\end{equation}
where $\mathcal{L}_{\mathcal{C}}$ is the loss function of classifier $C$. 

As a whole, the total objective is:
\begin{align}
    J_1 &= J_{\mathcal{C}}+\lambda _1 J_{\mathcal{D}},
\end{align}
where $\lambda_1$ is a hyper parameter to trade off the objectives. 

In the following, we will demonstrate the second stage of the knowledge transferring process, which is implemented by the two different mechanisms.
\subsection{The first mechanism: \oursstl}\label{fist}
\begin{figure}
  \centering
  \centerline{\includegraphics[scale=0.3]{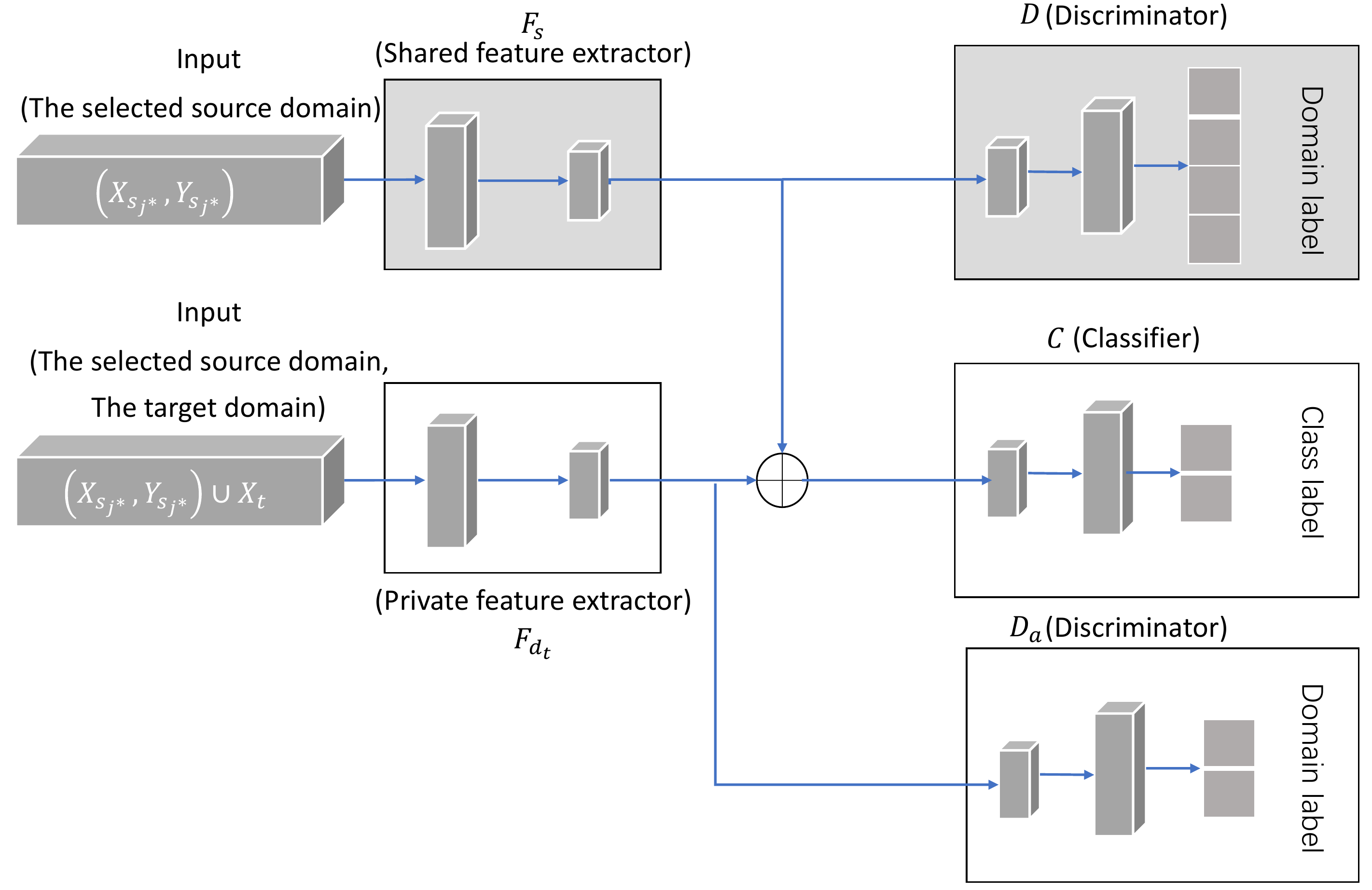}}
  \caption{Illustration of the selective domain adaptation (\oursstl) method. Parameters of $\mathcal{F}_s$, $\mathcal{F}_{d_j}$, $\mathcal{D}$ and $\mathcal{C}$ are pre-trained in the first stage. The target private extractor $\mathcal{F}_{d_t}$ is initialized according to the selected source domain. We utilize $\mathcal{D}_a$ to align the private feature distributions between the target domain and the selected source domain and transfer knowledge from the selected source domain. The parameters of the modules in the gray rectangles are fixed during the training process.}
  \label{step2}
\end{figure}

It is intuitive that not all knowledge from the source domains can benefit the target domain. In this subsection, we will introduce the selective domain adaptation (\oursstl) mechanism to selectively transfer the private knowledge from the closest source domain at the feature level. In detail, we first utilize an unsupervised metric named $\mathcal{A}$-distance to measure the intimacies (i.e. domain distances) between the source domains and the target domain and then transfer private knowledge from the closest source domain to the unlabeled target domain. As depicted in Figure~\ref{step2}, a particular discriminator $\mathcal{D}_a$ is employed to align the private feature distributions between the selected source domain and the target domain. Besides, we impose a parameter constraint to boost the performance of our learning system. At last, a private target extractor $\mathcal{F}_t$ for the target domain can be obtained. In the following, we interpret them in detail successively. 

\subsubsection{Measuring domain distances}
Intuitively, closer the source domain is to the target domain, better the domain adaptation performance will be. The $\mathcal{A}$-distance is verified as an effective metric to measure distances between domains~\cite{blitzer2007biographies,ben2007analysis}, based on which we will select the source domain(s) to transfer from in this paper. In principle, $\mathcal{A}$-distance measures domain adaptability by using the divergence of two domains. In practice, computing the $\mathcal{A}$-distance for a finite sample is exactly the problem of minimizing the empirical risk of a classifier that discriminates between instances drawn from $D_i$ and instances drawn from $D_j$~\cite{blitzer2007biographies}. This is convenient for us since it allows us to implement it in an unsupervised way. The measuring process can be divided into four steps: (1) assign domain label for each example and mix the two domains of data; (2) train a classifier on these merged data; (3) measure the classifier's $error$ on a held-out test set; and (4) calculate the $\mathcal{A}$-distance according to:
\begin{equation}
    {dis}_\mathcal{A}(D_i,D_j)=2(1-2 \min_{\substack{h \in\mathcal{H}}} error(h|D_i,D_j)),
\label{eq1}    
\end{equation}
where $\mathcal{H}$ is a hypothesis space and $h$ is the optimal proxy classifier that discriminates examples from which domain. 
In this paper, we choose a linear bag-of-words SVM as the proxy classifier to estimate the $error(h|D_i,D_j))$~\footnote{https://github.com/rpryzant/proxy-a-distance}. The smaller the distance is, the closer two datasets are. 

After the distance measuring, we select the closest source domain accordding to:
\begin{equation}
        j^* = \underset{j}{\arg \min}\ {dis}_\mathcal{A}(D_\mathcal{T},D_j).
\end{equation}
\subsubsection{Selective domain adaptation}
Then, we utilize $\mathcal{D}_a$ to align the private feature distributions between the selected source domain and the target domain and transfer private knowledge from the closest source domain through adversarial training, as shown in Figure~\ref{step2}. In detail, we first initialize the parameters of $F_t$ with the wights of the selected source-domain parameters and then update them through the adversarial training. The objective of $\mathcal{D}_a$ is:
\begin{equation}
        J_{\mathcal{D}_a}= \underset{x_i \sim  (\mathcal{S} _{j^*} \cup \mathcal{T})}{\mathbb{E}}\left[\mathcal{L}_{\mathcal{D}_a}\left(\mathcal{D}_a\left(\mathcal{F}_{t}(x_i)\right) ; d_i \right)\right],
\label{fd}
\end{equation}
where $\mathcal{T}$ represents the target domain, $j^*$ denotes the closest domain to the target domain, $\mathcal{L}_{\mathcal{D}_a}$ is the loss function of $\mathcal{D}_a$, and $d_i$ is the domain label for each instance. It needs to be noted that the target private extractor $\mathcal{F}_{t}$ has two functions: one is to help $\mathcal{C}$ perform better and the other is to confuse $\mathcal{D}_a$. 

It has been confirmed that if the weights of the source domain and the target domain are related but not shared can obtain superior performance~\cite{rozantsev2018beyond}, so we introduce the parameter constraint into our system to boost the performance. 
\subsubsection{Regularizing with parameter constraint}
The parameter constraint is imposed to ensure the parameters of the target domain have no great difference to the parameters of the source  domains. We write the objective of the parameter constraint as:
\begin{align}
    J_{\theta} &=\left \| \theta_s-\theta_t \right \|_2^2,
\end{align}
where $\theta_{s}$ corresponds to parameters of a source domain $\mathcal{F}_{d_{j^*}}$ and $\theta_{t}$ corresponds to parameters of $\mathcal{F}_{t}$.

\subsubsection{The training process}
\begin{algorithm}[htbp]
\caption{Selective Domain Adaptation (\oursstl) }
\label{alg1}
\textbf{Require}: $\left\{\mathcal{S}_{j}\right\}_{j=1}^{K}$, $\mathcal{T}$, hyperparameter $\lambda_2, \lambda_{\theta}$, $b\in\mathbb{N}$ (batch size), parameters of the pre-trained model\\
\begin{algorithmic}[1] 
\STATE Load corresponding parameters
\STATE $\triangleright \mathcal{D}_a$ iterations
\FOR {$m = 1$\ \textbf{to} \ $iter1$ }        
\STATE{$\ell _{\mathcal{D}_a} =0$} 
\STATE{Sample: $\mathbf{x_s} \triangleq (\boldsymbol{x}_{i}, \boldsymbol{d}_{i})_{i=1}^{b} \sim \mathcal{S}_{j^*}$ \hfill{$\triangleright$ Sample from the closest source domain}}
\STATE{Sample: $\mathbf{x_t} \triangleq (\boldsymbol{x}_{i}, \boldsymbol{d}_{i})_{i=1}^{b} \sim \mathcal{T}$}
\STATE{$\mathbf{f_d}=\mathcal{F}_{t}(\mathbf{x_s})$} 
\STATE{$\mathbf{f_t}=\mathcal{F}_{t}(\mathbf{x_t})$}
\STATE{$\ell _{\mathcal{D}_a}\mathrel{+}= J_{\mathcal{D}_a}(\mathcal{D}_a(\mathbf{f_d};d))+J_{\mathcal{D}_a}(\mathcal{D}_a(\mathbf{f_t};d)$}) \hfill{$\triangleright$ Equation (\ref{fd}})
\STATE{Update $\mathcal{D}_a$ parameters using $\nabla  \ell _{\mathcal{D}_a}$}
\ENDFOR
\STATE $\triangleright$Main  iterations
\FOR {$m = 1$\ \textbf{to} \ $iter2$ }        
\STATE $loss=0$
\STATE{Sample: $\mathbf{x_s} \triangleq (\boldsymbol{x}_{i}, \boldsymbol{y}_{i})_{i=1}^{b} \sim \mathcal{S}_{j^*}$}
\STATE{$\mathbf{f_s}=\mathcal{F}_s(\mathbf{x_s})$} 
\STATE{$\mathbf{f_d}=\mathcal{F}_t(\mathbf{x_s})$}
\STATE{$loss\mathrel{+}= J_{\mathcal{C}_{1} }(\mathcal{C}(\mathbf{f_s},\mathbf{f_d});\mathbf{y_s})$} \hfill{$\triangleright$ Equation (\ref{jc1}})
\STATE{Sample: $\mathbf{x_s} \triangleq (\boldsymbol{x}_{i}, \boldsymbol{d}_{i})_{i=1}^{b} \sim \mathcal{S}_{j^*}$ }
\STATE{Sample: $\mathbf{x_t} \triangleq (\boldsymbol{x}_{i}, \boldsymbol{d}_{i})_{i=1}^{b} \sim \mathcal{T}$}
\STATE{$\mathbf{f_d}=\mathcal{F}_t(\mathbf{x_s})$}
\STATE{$\mathbf{f_t}=\mathcal{F}_t(\mathbf{x_t})$}
\STATE{$loss\mathrel{+}=$ $ \lambda _2 \cdot \left ( J_{\mathcal{D}_a}(\mathcal{D}_a(\mathbf{f_d};d))+J_{\mathcal{D}_a}(\mathcal{D}_a(\mathbf{f_t};d) \right )$})
\STATE{$loss\mathrel{+}= \lambda_{\theta } J_{\theta}$} \hfill{$\triangleright$ Equation (\ref{jc2}})
\STATE{Update $\mathcal{F}_{t},\mathcal{C}$ parameters using $\nabla loss$}
\ENDFOR
\end{algorithmic}
\end{algorithm}
Finally, the private features and shared features of the target domain are concatenated and sent to the classifier $C$ to infer the comments' polarities:
\begin{equation}\label{jc1}
    J_{\mathcal{C}_1}= \sum_{i=1}^{\left | \mathcal{S}_{j^*} \right |} \underset{(x_i,y_i) \sim \mathcal{S}_{j^*} }{\mathbb{E}}\left[\mathcal{L}_{\mathcal{C}_1}\left(\mathcal{C}\left(\mathcal{F}_{s}(x_i), \mathcal{F}_{t}(x_i)\right) ; y_i\right)\right],
\end{equation}
where $\mathcal{L}_{C_1}$ is a NLL loss function. 

In sum, the total objective function of \oursstl \ can be written as below:
\begin{align}
    J_2 &= J_{\mathcal{C}_1}+\lambda _2 J_{\mathcal{D}_a}+\lambda_\theta{J_\theta}, 
\label{jc2}
\end{align}
where $\lambda_2$ and $\lambda_\theta$ are hyperparameters to balance each objective.

The whole \oursstl \ \ algorithm is summarized in Algorithm~\ref{alg1}. In detail, we first initialize the parameters of the target private extractor with the parameters of the closest source domain (line 1) and then train the discriminator $D_a$ (line 2 to line 9), followed by updating weights according to the Equation \ref{jc2}. The modules $\mathcal{F}_s$, $\mathcal{F}_t$, $\mathcal{C}$, $\mathcal{D}$, and $\mathcal{D}_a$ can be any suitable networks, such as CNNs (Convolutional Neural Networks), Bi-LSTM (bidirectional Long Short Term Memory networks), or MLP (Multi-Layer Perceptron). 

\subsection{The second mechanism: \ourssfe}\label{second}
\begin{figure}
  \centering
  \centerline{\includegraphics[scale=0.3]{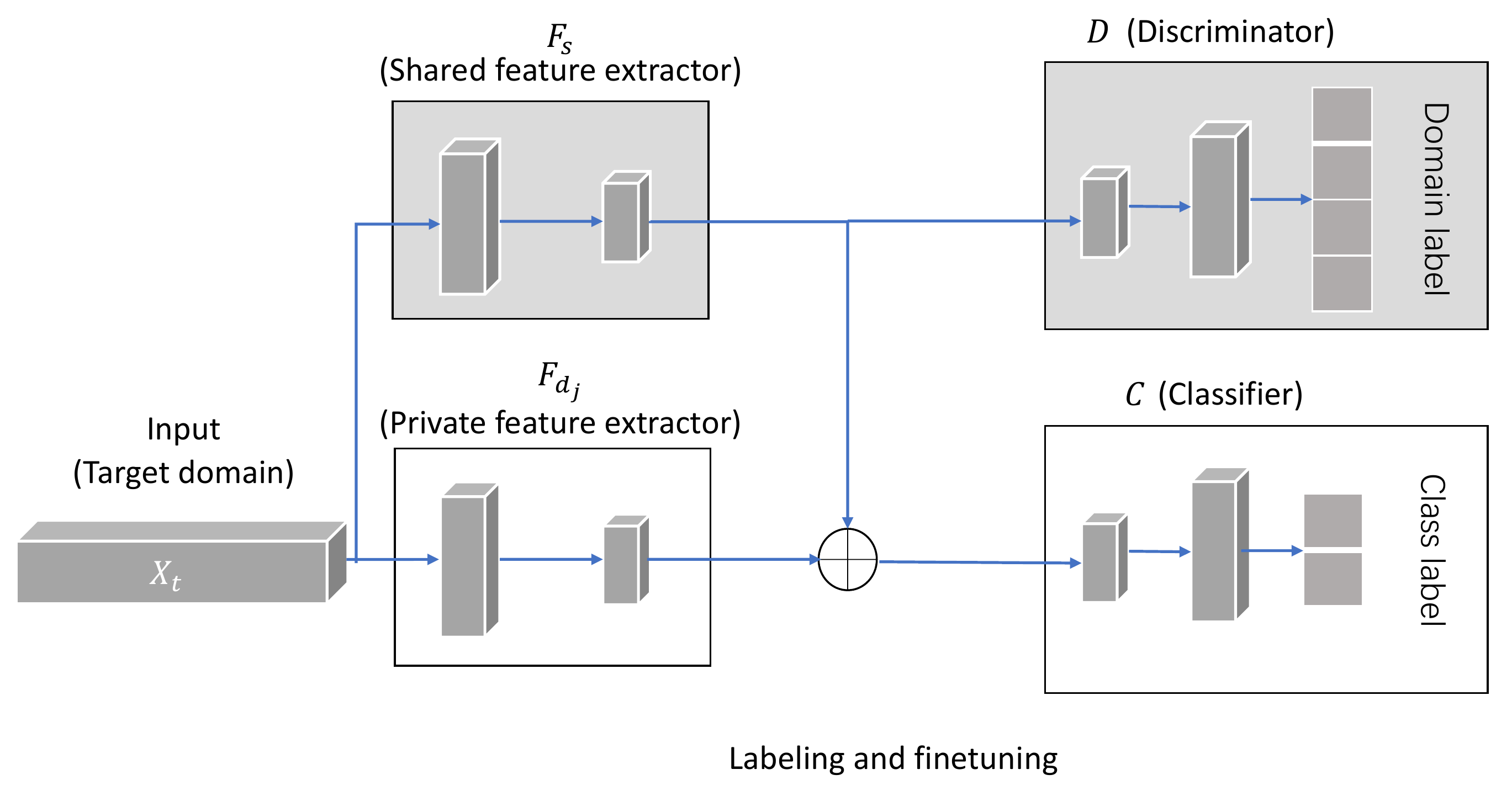}}
  \caption{Illustration of the target-oriented ensemble (\ourssfe) method. Parameters of $\mathcal{F}_s$, $\left \{ \mathcal{F}_{d_j} \right \}_{j=1}^{3}$, $\mathcal{D}$ and $\mathcal{C}$ are pre-trained in the first stage. Pseudo labels are obtained firstly and then utilized to update the parameters of $C$ and $\mathcal{F}_{d_j}$.
  The parameters of the modules in the gray rectangles are fixed during the training process.}
  \label{toe}
\end{figure}

Motivated by the Tri-training~\cite{zhou2005tri} method and ensemble learning~\cite{zhang2012ensemble,polikar2012ensemble} methods, we propose another target-oriented ensemble (\ourssfe) mechanism, which is illustrated in Figure~\ref{toe}. Take the merit of ensemble learning, we first obtain some confident pseudo labels for the unlabeled target instances annotating by the top-3 closest source domains. Then, the selected source-domain extractors are finetuned based on these labeled examples. This mechanism has two advantages: (1) integratedly annotated by the top three closest source domains, the pseudo labels are more consistent and confident; (2) finetuned by the target domain pseudo labels, the final decision boundary will be more oriented to the target domain.

The detailed training process is described in Algorithm~\ref{alg2}. As a whole, we first choose the top-3 closest classifiers ($\left \{ \mathcal{F}_{d_j} \right \}_{j=1}^{3}$) (line 1 to 3) by:
\begin{equation}
        j \in  \underset{j}{\arg \operatorname{min\_3}}\ {dis}_\mathcal{A}(D_\mathcal{T},D_j),
\end{equation} 
where $\operatorname{min\_3}$ means the three smallest distances between the target domain and the source domains. Then, these selected source domain classifiers are used to annotate a set of pseudo labels $\mathcal{T}_l$ (line 4 to 11). Next, $\left \{ \mathcal{F}_{d_j} \right \}_{j=1}^{3}$ are finetuned with $\mathcal{T}_l$ (line 13 to 16) by:
\begin{equation}
    J_{\mathcal{C}_2}= \sum_{j=1}^{3}\sum_{i=1}^{\left | \mathcal{T}_l \right |} \underset{(x_i,y_i) \sim \mathcal{T}_l }{\mathbb{E}}\left[\mathcal{L}_{\mathcal{C}_2}\left(\mathcal{C}\left(\mathcal{F}_{s}(x_i), \mathcal{F}_{d_j}(x_i)\right) ; y_i\right)\right],
\label{j2}
\end{equation}
where $\mathcal{L}_{C_2}$ is a NLL loss function. At last, the finetuned classifiers are exploited to make inference for the target instances by averagely assembling the results.
\begin{algorithm}[htbp]
\caption{Target-oriented ensemble (\ourssfe) }
\label{alg2}
\begin{algorithmic}[1] 
\REQUIRE{$ \mathcal{T}$, parameters of pre-trained model, $b\in\mathbb{N}$ (batich size)}
\STATE{Select the top-3 closest source domains}
\STATE{Load parameters: $\mathcal{F}_s, \left \{ \mathcal{F}_{d_j} \right \}_{j=1}^{3}, D, C$} 
\STATE{Initialize: $ \eta =0.02, \Delta=0.98, N=10, \mathcal{T}_l = \varnothing, \mathcal{L} = \varnothing$}
\STATE $\triangleright labeling$ iterations
\WHILE{$\left | \tau^{-1} \right | + \left | \tau^{-2}  \right | \geqslant  N \ or \ \Delta \geqslant  0.5$}
\STATE{Sample: $\boldsymbol{x}_t \triangleq (\boldsymbol{x}_{i},\boldsymbol{d}_{i})_{i=1}^{b} \sim \mathcal{T}$}
\STATE{$\boldsymbol{\hat{y}}_{i}= labeling(\mathcal{F}_{s}\left(\boldsymbol{x}_{t}\right), \mathcal{F}_{d_{j}}\left(\boldsymbol{x}_{t}\right))$}
\STATE{$\mathcal{L}\triangleq\left(\boldsymbol{x}_{i}, \hat{\boldsymbol{y}}_{i}\right)$}
\STATE{$\mathcal{T}_l=\mathcal{T}_l\cup \mathcal{L}$}
\STATE{$\mathcal{T} = \mathcal{T} \ \backslash \ \mathcal{L}$}
\STATE{$\Delta =\Delta -\eta $}
\ENDWHILE
\STATE $\triangleright Finetuning$ iterations
\FOR {$k = 1$\ \textbf{to} \ $iter$ }        
\STATE{Sample: $\boldsymbol{\hat{x}}_{t} \triangleq (\boldsymbol{\hat{x}}_{i}, \boldsymbol{\hat{y}}_{i})_{i=1}^{b} \sim \mathcal{T}_l$}
\STATE{$loss = \mathcal{L}_{\mathcal{C}_2}\left(\mathcal{C}\left(\mathcal{F}_{s}(\hat{x}_{t}), \mathcal{F}_{d_j}(\hat{x}_{t})\right)\right)$ \hfill{$\triangleright$ Equation~\ref{j2}}}
\STATE{Update $\mathcal{F}_{d_j},C$ by using $\nabla loss$}
\ENDFOR
\end{algorithmic}
\end{algorithm}
For more details, to derive more confident labels we just trust the annotated labels with a high accuracy possibility at the very beginning through the adoption of a dynamic possibility threshold $\Delta$ (e.g. 0.98) and a gradually decreasing constant $\eta =0.02$. In final, the number of increasing labels (denoted as $N$) of $\mathcal{T}_l$ between two successive iterations (denoted as $\tau^{-1}$ and $\tau^{-2}$) would very small (e.g., 10) and at this time we terminate the labeling. In addition, we denote $iter$ as the iteration of training, and the function $labeling$ means the method of labeling by source classifiers. 

\section{Experiment}
In this section, we investigate the empirical performance of our proposed framework on related sentiment classification tasks and then compare our framework to other state-of-the-art models. As a whole, we use a 4-domain dataset (i.e. \textbf{Amazon review dataset}) to validate the effectiveness of $\mathcal{A}$-distance and the first mechanism \ \oursstl \ and a 16-domain dataset (i.e.,\textbf{FDU-MTL}) to comprehensively verify our framework. In the following, we first introduce the experiment setup (\ref{setup}), then describe the diverse experiments on the two datasets separately (\ref{e1} and \ref{e2}), and at last dive into the problems presented in two experiments (\ref{dis}).

\subsection{Experimental Setup}\label{setup}
\paragraph{Datasets}
We choose two datasets for different text classification tasks, which are widely used in multi-domain text classification tasks. They are briefly described as follows:
\begin{itemize}
    \item $\textbf{Amazon review dataset}$~\cite{blitzer2007biographies} contains 2000 samples for each of the four domains: book, DVD, electronics, and kitchen with binary labels (i.e. positive, negative). The features of this dataset are already pre-processed into 30000-dimensional vectors.
    \item $\textbf{FDU-MTL}$~\cite{liu2017adversarial} includes reviews from 14 Amazon domains: books, electronics, DVD, kitchen, apparel, camera, health, music, toys, video, baby, magazine, software, and sports, and two movie reviews from the IMDb and the MR dataset. Each domain has a development set of 200 samples and a test set of 400 samples. The number of training data and unlabeled data vary across domains but are roughly from 1400 to 2000.
\end{itemize}
\paragraph{Competitor Methods for MS-UDA}
Competitor methods on \textbf{Amazon review dataset} include:
\begin{itemize}
    \item \textbf{mSDA}~\cite{chen2012marginalized}: it utilizes marginalized stacked denoising autoencoders to learn new representations for domain adaptation computationally and can be computed in closed-form.
    \item \textbf{DANN}~\cite{ganin2016domain}: it introduces an adversarial training-based approach to learn the domain invariant representation. 
    \item \textbf{MDAN(H-MAX), MDAN(S-MAX)}~\cite{zhao2017multiple}: they are two adversarial neural models: the first model optimizes directly their proposed bound (H-MAX), while the second model is a smoothed approximation of the first one (S-MAX), which is more data-efficient and task-adaptive.
\end{itemize}
Competitor methods on \textbf{FDU-MTL} are briefly described as follows:
\begin{itemize}
    \item $\textbf{ASP-MTL-SC, ASP-MTL-BC}$~\cite{liu2017adversarial}: these two models are single-channel model and bi-channel model of adversarial multi-task learning.
    \item $\textbf{MAN}$~\cite{chen2018multinomial}: this model is a multinomial adversarial network for multi-domain text classification.
    \item \textbf{Meta-MTL}~\cite{chen2018meta}: this model combines the multi-task learning and meta-learning to capture the meta-knowledge of semantic composition and generate the parameters of the task-specific semantic composition models.
    \item \textbf{BERT-base}~\cite{devlin2018bert}: it is a widely adopted pre-trained model and a strong baseline for sentiment analysis.
    \item \textbf{DistilBERT-base}~\cite{sanh2019distilbert}: it is a distilled version of BERT, which is powerful but smaller, faster , and cheaper than \textbf{Bert-base}.
\end{itemize}
\paragraph{Implement details}
We adopt the WGAN~\cite{arjovsky2017wasserstein} training strategy, which is more stable and less sensitive to model architecture and choice of hyperparameter configurations, to implement the adversarial training in our model. The WGAN training tricks including: (1) training the system in two steps as a whole: firstly fix the parameters of the classifier and train the discriminator, then fix the parameters of the discriminator and train the classifier; (2) use Wasserstein loss to train the discriminator and encoder, which can be seen in Equation~\ref{eqa:d}; (3) update the discriminator more times than the generator each iteration (e.g. 5).

For a fair comparison, we follow the setting of \textbf{MAN} and choose word2vec~\cite{mikolov2013efficient} to initialize our word embedding and set shared features 128d and private features 64d. We adopt Adam~\cite{kingma2014adam} as the optimizer, set the learning rate 0.0001, the dropout rate 0.4, and batch size 16. 

\subsection{Experiments on \textbf{Amazon review dataset}}\label{e1}
For a fair comparison, we follow most of the settings in~\cite{chen2018multinomial}, which including : (1) employ MLPs as feature extractors; (2) cut out 5000 most frequent features of each view as a 5000d input feature vector; (3) take turns choosing one domain as the target domain and the rest as the source domains.
\subsubsection{The effectiveness of $\mathcal{A}$-distance}
\begin{table}[htbp]
\caption{$D_s$ and $D_t$ denote the source domain and the target domain respectively. Each result is adapted from books, dvd, electronics ,and kitchen (except itself) respectively and the underlined result is the best result.}
\centering
\begin{tabular}{ccccc}
\hline
\diagbox{$D_s$}{$D_t$}&books&dvd&electronics&kitchen\\ 
\hline
books& -& 78.32& \uline{78.68}& 78.30\\
dvd& \uline{81.23}& -& 80.70& 79.98\\
electronics& 84.32& 84.05& -& \uline{85.06}\\
kitchen& 86.01& 85.63& \uline{87.33}& -\\
\hline
\end{tabular}

\label{tabt}
\end{table}
\begin{figure}
  \centering
  \includegraphics[scale=0.08]{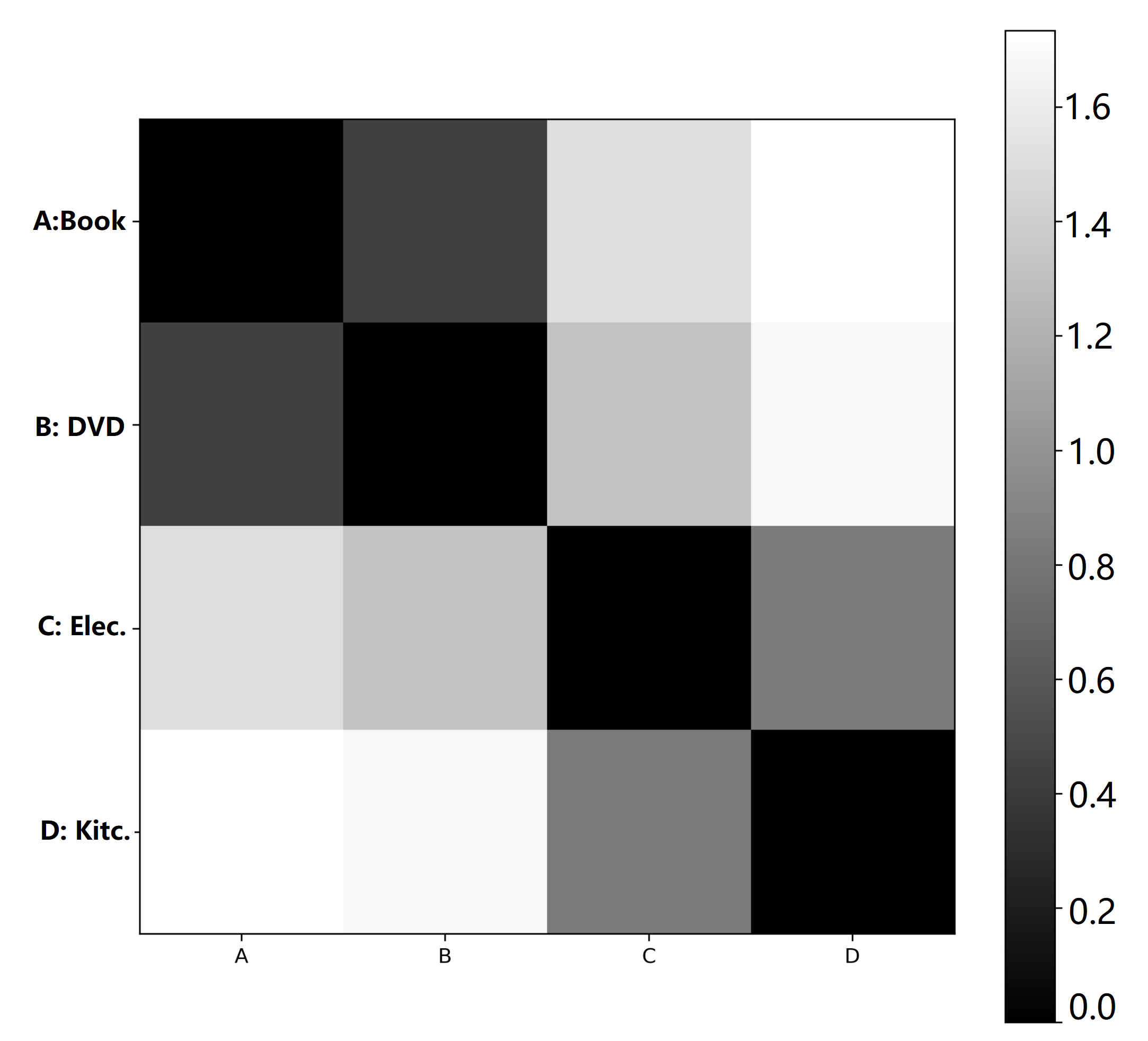}\\
  \caption{The heatmap of $\mathcal{A}$-distance on Amazon review dataset. Lower numbers or darker colors mean closer to each other. From A to D represent book, DVD, electronics, kitchen respectively.}\label{A-distance1}
\end{figure}
In order to validate the effectiveness of $\mathcal{A}$-distance, we conduct experiments transferring from different source domains by \oursstl. The $\mathcal{A}$-distances between domains are pictured in Figure \ref{A-distance1} and the corresponding results are demonstrated in Table~\ref{tabt}. These results are obtained by transferring private knowledge from one source domain to the target domain rather than setting private features to zeroes like in~\cite{chen2018multinomial}. From Table \ref{tabt} we can see that the adaptation performance will be better if the $\mathcal{A}$-distance between the source domain and target domain is smaller.
\subsubsection{The performance of \oursstl}
\begin{table}
\caption{Comparative results on the Amazon review dataset. Both the private feature and shared feature is 128d. Except for our model the rest is taken from~\cite{chen2018multinomial}. $MDAN^1$ and $MDAN^2$ represent H-MAX and S-MAX respectively. $MAN^1$ and $MAN^2$ represent MAN-MLP with L2 loss and NLL loss respectively. Best results are underlined.}
\centering
\begin{tabular}{l|cccc|c}
\hline
Target Domain  & Books & DVD& Elec. & Kit  & \textbf{Avg} \\
\hline
MLP & 76.55  & 75.88 & 84.60 & 85.45 & 80.46     \\
mSDA & 76.98 & 78.61 & 81.98 & 84.26 & 80.46      \\
DANN & 77.89 & 78.86 & 84.91 & 86.39 & 82.01 \\
$MDAN^1$ & 78.45 & 77.97 & 84.83 & 85.80 & 81.76 \\
$MDAN^2$ & 78.63 & 80.65 & \uline{85.34} & 86.26 & 82.72 \\
$MAN^1$ & 78.45 & 81.57 & 83.37 & 85.57 & 82.24 \\
$MAN^2$ & 77.78 & \uline{82.74} & 83.75 & 86.41 & 82.67     \\
\oursstl & \uline{78.68} & 81.23 & 85.06 & \uline{87.33} & \uline{83.08}     \\
\hline
\end{tabular}
\label{tab1}
\end{table}

The results comparing with other baseline methods are demonstrated in Table \ref{tab1}. Due to the features are ready-made and the dimension of private features is comparably small, the performance we can boost is limited. However, there is also 0.41 percent advance on average can be observed. So the experimental results can roughly demonstrate the importance of the private features and further verification will be conducted on the \textbf{FDU-MTL} dataset.

\subsection{Experiments on \textbf{FDU-MTL}}\label{e2}
\begin{table}[htbp]
  \centering
  \caption{Results on the FDU-MTL dataset. The best results are underlined. For column \textbf{SDA}, the domains in brackets denote the selected source domains. The numbers in brackets in the last row are the comparative performance with the best average accuracy. The columns \textbf{A-Ens., L-Ens., T-Ens.} stand for the traditional average ensemble method, the last-3 ensemble method, the top-3 ensemble method.}
  \setlength{\tabcolsep}{0.5mm}{}
    \begin{tabular}{l|cccccc|c|cccc}
    \toprule
    Target & $\text{ASP}^1$  & $\text{ASP}^2$  & Meta  & Man   & BERT  & Distil & SDA   & A-Ens. & L-Ens. & T-Ens. & TOE \\
    \midrule
    books & 83.2  & 83.7  & 86.3  & 86.3  & 81.8  & 84.3  & 87.7(toys) & 85.6  & \uline{88.5}  & 87.8  & 87.8 \\
    electronics & 82.2  & 83.2  & 86.0    & 88.0    & 88.3  & 83.5  & \uline{89.0}(kit.) & 85.8  & 86.2  & 88.0    & 88.0 \\
    dvd   & 85.5  & 85.7  & 86.5  & 87.5  & 82.5  & 83.0    & 87.9(mag.) & 88.0    & 87.7  & 87.8  & \uline{88.3} \\
    kitchen & 83.7  & 85.0   & 86.3  & 89.3  & 86.3  & 84.5  & \uline{89.8}(cam.) & 87.5  & 89.5  & 89.3  & 89.0 \\
    apparel & 87.5  & 86.2  & 86.0    & 86.5  & 84.8  & 81.3  & \uline{87.6}(mus.) & 87.5  & 85.5  & 86.8  & 86.0 \\
    camera & 88.2  & \uline{89.7}  & 87.0    & 86.5  & 87.3  & 83.0 & 87.4(toys) & 87.6  & 85.7  & 87.5  & 87.3 \\
    health & 87.7  & 86.5  & 88.7  & 88.0 & \uline{91.8}  & 85.5  & 89.1(baby) & 86.2  & 87.2  & 88.5  & 88.3 \\
    music & 82.5  & 81.7  & 85.7  & 85.8  & 80.0   & 82.8  & 86.2(mag.) & 83.4  & 81.7  & 82.5  & \uline{86.8} \\
    toys  & 87.0& 88.2  & 85.3  & 88.5  & 88.5  & 88.0   & 89.3(baby) & 87.7  & 88.0    & 90.0  & \uline{90.3} \\
    video & 85.2  & 85.2  & 85.5  & 85.5  & 81.0    & 78.5  & \uline{87.5}(sof.) & 86.1  & 84.2  & 85.5  & 87.3 \\
    baby  & 86.5  & 88.0   & 86.0    & 88.0    & 85.8  & 85.5  & \uline{88.6}(spo.) & 86.1  & 87.5  & 88.3  & 88.3 \\
    magazines & \uline{91.2}  & 90.5  & 90.3  & 83.0 & 82.5  & 83.5  & 84.1(vid.) & 84.7  & 85.7  & 85.0   & 85.8 \\
    software & 85.5  & 88.2  & 86.5  & 85.0 & 87.5  & 76.8  & 85.7(IMDB) & \uline{88.8}  & 84.3  & 85.8  & \uline{88.8} \\
    sports & 86.7  & 86.5  & 85.7  & 86.5  & 86.5  & 85.8  & 87.2(baby) & \uline{87.8}  & 87.5  & \uline{87.8}  & \uline{87.8} \\
    IMDB  & 87.5  & 86.7  & 87.3  & 84.3  & 80.8  & 74.3  & 84.8(mus.) & 87.8  & \uline{88.0}    & 87.8  & 87.8 \\
    MR    & 75.2  & \uline{76.5 } & 75.5  & 76.3  & 74.5  & 68.8  & 76.4(sof.) & \uline{76.5} & 74.5  & 74.0& 74.3 \\
    \midrule
    Avg.   & 85.3  & 85.7  & 85.9  & 85.9  & 84.4  & 81.8  & 86.8(+0.9) & 86.1(0.2) & 85.7(-0.2) & 86.4(0.5) & \uline{87.0}(1.1) \\
    \bottomrule
    \end{tabular}
  \label{tab3}%
\end{table}%

For experiments on \textbf{FDU-MTL}, we employ CNNs as feature extractors with ReLU as the activation function. We take turns choosing 15 tasks as source domains and the left one as target domain and then transfer private knowledge to the unlabeled target domain. The results from our competitors and our proposed framework are presented in Table \ref{tab3}. The results of columns named Bert and Distil (i.e. DistilBert) are implemented based on the open source project simpletransformer~\footnote{https://github.com/ThilinaRajapakse/simpletransformers}. The columns named \oursstl \ and \ourssfe \ denote the results of our proposed framework using the corresponding mechanism. The domain names, which are abbreviated in brackets of column \oursstl, denote the selected source domains. Specially, to demonstrate the effectiveness of our elaborately designed ensemble mechanism \ourssfe, we conduct another 3 series of experiments named \textbf{A-Ens., L-Ens., T-Ens.}. Among them, \textbf{A-Ens} denotes the average ensemble method by averaging all the source-domain results; \textbf{L-Ens.} and \textbf{T-Ens.} denote the ensemble methods by averaging the results of 3-farthest and 3-closest source domains separately.
\subsubsection{The importance of the private features}
To further demonstrate the importance of private features, we set all the private features to zeros and reimplement the \textbf{MAN}\footnote{https://github.com/ccsasuke/man}. The accuracy of 85.9 with all zero private features is already a good result, which proves the effectiveness of the basic share-private model. Moreover, the obvious further boosts (i.e. 0.9 and 1.1 on average respectively) are obtained by adopting our proposed framework with \oursstl \ and \ourssfe, which demonstrates the importance of the private features and the effectiveness of our proposed mechanisms.
\subsubsection{The performance of \oursstl \ and \ourssfe}
For \oursstl, we get 1.5, 1.1 and 0.9 percent improvement on average comparing with $\textbf{ASP-MTL-SC}$, $\textbf{ASP-MTL-BC}$ and \textbf{MAN} respectively.

Looking at the series of the experiments for \ourssfe, we can conclude from Table~\ref{tab3} that: (1) the traditional ensemble method (i.e. \textbf{A-Ens.}) obtains a slight boost (i.e. 0.2 on average) to \textbf{MAN}; (2) the accuracy of \textbf{L-Ens.} drops 0.2 on average, which implies if we select the transferred source domains not so well the ensemble performance may be worse; (3) the accuracy of the \textbf{T-Ens.} is 0.5 percentage higher than \textbf{MAN}, which tells that if the source domains are closer to the target domain, the ensemble performance will be better; (4) our proposed \textbf{TOE} acquire 1.1 and 0.9 percentage boost compared to the best baseline method \textbf{MAN} and the average ensemble method \textbf{A-Ens.} separately, which demonstrates the effectiveness of the second mechanism.
\subsection{Discussion}\label{dis}
To further verify the effectiveness of our framework, we compare our framework with the supervised state-of-the-art results method $\textbf{MAN}$~\cite{chen2018multinomial}. The performance of \oursstl \ and \ourssfe \ outperforms MAN on 4 domains and 7 domains of all the 16 domains separately. On average, the accuracies are 1.6 and 1.1 percent lower than the supervised $\textbf{MAN}$, so there is room for us to bridge this gap for future work.

For source domain selection in the domain adaptation, we can draw the following conclusions: (1) for experiments on \textbf{Amazon review}, selecting a close source domain to transfer from is very important. For example, transferring from 'books' to 'dvd' is 1.25 percentage better than transferring from 'kitchen'; (2) for experiments on \textbf{FDU-MTL}, source domains at least could be divided into two clusters: one is close to target domain and one is remote to the target domain. If we transfer knowledge from any source domain coming from the close cluster, the performance would vary a little (e.g. less than 0.5 percentage of accuracy), but if we transfer from one of the source domains coming from the remote cluster, the performance will degrade a lot (e.g. more than 1 percentage of accuracy); 
(3) exposure to too many source domains for ensemble learning without considering the domain relevance is detrimental to the target domain performance, which is also can be seen in~\cite{gururangan2020don};
(4) we further believe that along with the increase of dimension of the private features, our method will perform much better than other methods because our proposed method can transfer the private knowledge from other source domains effectively.

\section{Conclusion and Future Work}
In this paper, we propose a two-stage MTL framework, through which we can implement the unsupervised sentiment analysis with multiple source domains. To fulfill the knowledge transfer, we embed two mechanisms into the shared-private architecture. Experimental results show that our framework with the two novel mechanisms can improve the performances of a group of related tasks.

However, there is still a large room for the improvement of the domain selection. In future work, we would like to investigate how to select a source domain automatically and explore more effective methods to decide better dimensions of public and private features.

\section*{Compliance with Ethical Standards}

\textbf{Funding}
This work was funded by the National Key R\&D Program of China (No. 2018YFB1005100 \& No. 2018YFB1005104). 

\noindent\textbf{Conflict of Interest}
The authors declare that they have no conflict of interest.

\noindent\textbf{Ethical approval}
This article does not contain any studies with human participants or animals performed by any of the authors.

\noindent \textbf{Informed Consent}
Informed consent was not required as no humans or animals were involved.

\bibliographystyle{spmpsci}      
\bibliography{multi}   

\end{document}